# See360: Novel Panoramic View Interpolation

Zhi-Song Liu, *Member, IEEE*, Marie-Paule Cani, and Wan-Chi Siu, *Life Fellow, IEEE*

*Abstract*—We present See360, which is a versatile and efficient framework for 360° panoramic view interpolation using latent space viewpoint estimation. Most of the existing view rendering approaches only focus on indoor or synthetic 3D environments and render new views of small objects. In contrast, we suggest to tackle camera-centered view synthesis as a 2D affine transformation without using point clouds or depth maps, which enables an effective 360° panoramic scene exploration. Given a pair of reference images, the See360 model learns to render novel views by a proposed novel Multi-Scale Affine Transformer (MSAT), enabling the coarse-to-fine feature rendering. We also propose a Conditional Latent space AutoEncoder (C-LAE) to achieve view interpolation at any arbitrary angle. To show the versatility of our method, we introduce four training datasets, namely UrbanCity360, Archinterior360, HungHom360 and Lab360, which are collected from indoor and outdoor environments for both real and synthetic rendering. Experimental results show that the proposed method is generic enough to achieve real-time rendering of arbitrary views for all four datasets. In addition, our See360 model can be applied to view synthesis in the wild: with only a short extra training time (approximately 10 mins), and is able to render unknown real-world scenes. The superior performance of See360 opens up a promising direction for camera-centered view rendering and 360° panoramic view interpolation.

*Index Terms*— View rendering, 3D scene, adversarial network.

## I. Introduction

360° video/image, including panoramic, spherical or OmniDirectional Video (ODV), is a new type of multimedia that provides users with an immersive experience. For example, people can wear a Virtual Reality (VR) headset and move their heads to look around in a virtual world. The use of 360° video is then mandatory to achieve a real-time, realistic rendering of the scene. These techniques are useful in real-world applications, e.g., for displaying panoramic images/videos for vehicle driving, shopping, sightseeing and so on. In order to collect the training data, specific equipment such as Yi Halo and GoPro Odyssey, is generally required. We can also use RGB-D cameras to capture depth for 3-DoF or 6-DoF rendering, enabling depth estimation [1], [2], semantic segmentation [3]–[5] and salience prediction [6]–[8].

Another related topic is neural view rendering. The idea is to use deep learning approaches to explicitly [18]–[22] or implicitly [24]–[28] discover the 3D structure of objects for novel view synthesis [18], [22], [24] or neural rerendering [29]–[31]. In contrast with 360° video processing, novel view rendering usually takes an object as the view center. By pointing the camera toward it, different viewpoints of the same object are captured, possibly using a depth camera to get more geometric information. These views are then interpolated using a trained neural network. Many public synthetic datasets [32]–[34] are used for this task.

In contrast with both 360° video and novel view rendering, but bridging the gap between them, our goal is to achieve *camera centered, 360° panoramic novel view interpolation*. More precisely, our method allows using a single, ordinary camera to capture a few reference views of the real world and predict the intermediate views in between, achieving horizontal 360° view synthesis from this sparse input. Hence we refer to our proposed approach as See360. Note that as opposed to 360° video, we do not need any complex device to capture omnidirectional views for scene understanding. Our method also differs from novel view rendering since our goal is to capture the 3D structure of the surroundings rather than the structure of a single object. The problem we are tackling is thus more challenging. Since it is not possible to predict an unseen view without any prior, we use two references (left and right) views, to "interpolate" the intermediate views. The angle distance between references can be 60° or even up to 120° (note that the most common setting of field of view (FOV) for cameras is 55°, which ensures that the input views will overlap). All occluded and missing scene parts in between are estimated by our method, while achieving smooth view transition. In addition, we can achieve view interpolation for 360° panoramic videos/animations. Note that our proposed method is purely image-based view synthesis without requiring any knowledge of depth or 3D information, hence it is low cost on data requirement that common cameras can be used for view synthesis, but there might be spatial misalignment caused by the large moving objects, low scene overlapping or sudden scene changes.

Figure 1 illustrates our method: we attach a single camera to a electronic tripod head (Figure 1(a)) which is able to automatically rotate the camera over the full 360° range, to capture $T$ sparse reference views and a dense set of $S$ ($S > T$) intermediate views. For training, we place the camera at the

Manuscript received August 23, 2021; revised December 3, 2021; accepted January 18, 2022. Date of publication February 9, 2022; date of current version February 15, 2022. This work was supported in part by the Google Chair at École Polytechnique, The Hong Kong Polytechnic University, and the Caritas Institute of Higer Education, Hong Kong, SAR, under University Grants Committee (UGC), under Grant UGC/IDS(C)11/E01/20 (manpower) and Grant IDS(R)11/19 (high-performance computing facility). The associate editor coordinating the review of this manuscript and approving it for publication was Dr. Lu Fang. *(Corresponding author: Zhi-Song Liu.)*

Zhi-Song Liu was with CNRS (LIX), IP Paris, École Polytechnique, 91120 Palaiseau, France. He is now with the Caritas Institute of Higher Education, Hong Kong (e-mail: zhisong.liu@connect.polyu.hk).

Marie-Paule Cani is with CNRS (LIX), IP Paris, École Polytechnique, 91120 Palaiseau, France (e-mail: marie-paule.cani@polytechnique.edu).

Wan-Chi Siu is with the Caritas Institute of Higher Education, Hong Kong, and also with the Department of Electronic and Information Engineering, The Hong Kong Polytechnic University, Hong Kong (e-mail: enwcsiu@polyu.edu.hk).

This article has supplementary downloadable material available at https://doi.org/10.1109/TIP.2022.3148819, provided by the authors.

Digital Object Identifier 10.1109/TIP.2022.3148819

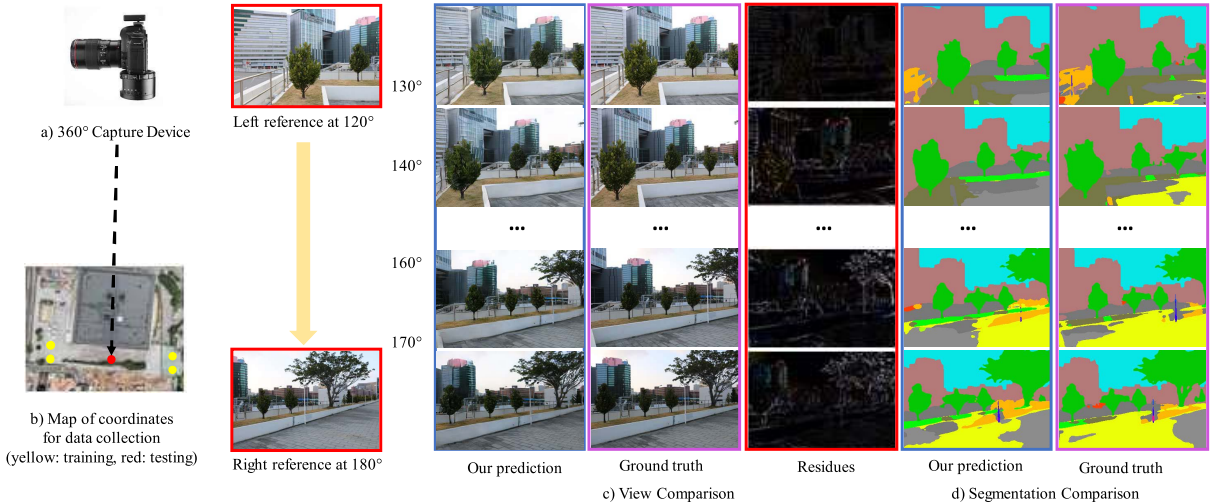

Fig. 1. (a) 360° capture device. (b) The coordinates where we place the camera for data collection. Given reference views at 120° and 180°, our proposed See360 model can render views at different angles from left to right. (c) Novel views comparison (include our prediction, ground truth and residues between them) and (d) Segmentation results comparison. The view changes from the left and right references can be observed from the buildings and trees.

four corners of a street to collect 4 such data-sets (at each location, we capture 6 images to cover 360° view), enabling the model to learn the 3D structure of the street. At the testing stage, we randomly place the camera in-between, for instance here at the center of the street (see, respectively, the red dot in Figure 1(b)). We then use $T=2$ views at 120° and 180° from this new position as references, to predict views from the whole angular range in between. As shown in Figure 1(c), the generated images seamlessly change along with the camera pose and are perceptually similar to ground truth. We also depict the residual difference between prediction and ground truth, magnified by 5 for better visualization. This shows that the very small differences are mainly located around edges, which indicates that the global information matches in the low frequency domain, despite of some high frequency information losses. This good pixel fidelity enables to use the generated images for other applications, such as semantic segmentation (see Figure 1(d)).

To summarize, we implicitly discover 3D correlations between reference views by using explicit 2D affine transformation to render novel views. Our key contributions are:

- To generate high-quality, photo-realistic images without requiring 3D information, we propose a Multi-Scale Affine Transformer MSAT) to render reference views in the feature domain using 2D affine transformation. Instead of learning one-shot affine transform to reference views, we learn multiple affine transforms in a coarse-to-fine manner to match the reference features for view synthesis.
- Furthermore, to allow users to interactively manipulate views at any angle, we introduce the Conditional Latent space AutoEncoder (C-LAE). It consists of 1) patch based correlation coefficients estimation and 2) conditional angle encoding. The former enables finding global features for 3D scene coding and the latter introduces target angles as one-hot binary codes for view interpolation.
- In addition, we provide two different types of datasets to train and test our model. One is the synthetic 360° images collected from the virtual world, including UrbanCity360 and Archinterior360. Another is the real 360° images collected from real-world indoor and outdoor scenes, including HungHom360 and Lab360. Semantic segmentation maps are also provided for all datasets. Our tests in the wild show that See360 can also be used, to some extent, with unknown real scenes. With a small number of training images (about 24 images) required, it takes 10 mins training to reconstruct the 360° view rendering.

## II. Related Work

In this section, we give a detailed review of previous related works on 1) 360° video processing, 2) neural view rendering and 3) 3D-aware view synthesis. Note that our work is also related to some classic image processing problems like image warping [59], [60], image stitching [61], [62] and inpainting [63], [64]. However, these methods do not study camera pose guided view interpolation, where our proposed method predict novel views given random camera poses.

### A. 360° Video/Image Processing

360° video/image has been increasingly popular and drawn great attention. With the available, commercial head mounted displays (HMDs), users can move freely their heads to have immersive experiences. New challenges recently raised for 360° video/image processing: 1) storage and transmission and 2) viewpoint-centric processing. For the first question, a video/image of very high resolution is required to achieve a good covering of the whole $360 \times 180°$ viewing range. Furthermore, to avoid viewers' motion sickness [10], a high frame rate is required. Many organizations have developed compression standards for 360° video/image such as MPEG-I [11] and JPEG-360 [12]. However, 360° video/image compression is still an ongoing research topic, since visual quality assessment (VQA) is needed to evaluate the degradation of compression. In real-world environments, the generation of 360° video/image follows equi-rectangular projection (ERP) corresponding to the spherical coordinate system.

Therefore, the resulting images need to be projected to the standard view for artifact-free visualization. With the development of deep learning, 360° videos have been well studied in many fields, such as depth estimation, or semantic segmentation. For example, Zioulis et al. [15] designed an autoencoder to model depth on omnidirectional imagery. Eder introduced plane-aware loss for dense depth estimation [17], as well as tangent transform to mitigate spherical distortion for segmentation [3]. Projecting images to the icosahedron spheres [16] is also another choice for 360° video/image processing.

*B. Neural View Rendering*

Neural view rendering is the creation of photo-realistic imagery of virtual worlds. Learning the 3D scene representation, rendering methods can render images for a variety of complex real-world phenomena. Let us focus on two applications: 1) neural rerendering and 2) novel view synthesis. Among neural rerendering methods, *Neural Rerendering in the Wild* [29] uses a neural network to synthesize realistic views of tourist landmarks with various lighting conditions. Pittaluga et al. [30] proposed to learn an invert reconstruction from point clouds to realistic novel views with unknown keypoint scale, orientation and multiple image sources. Neural avatar [31] is a deep network that produces body renderings with various body poses and camera positions. For novel view synthesis, the idea is to use multiple images or 3D models to render a new view of the object. [35]–[38], [67]–[69] propose generative adversarial networks for unsupervised learning of 3D representations. It can achieve random 3D pose rendering by a rigid-body transformation of the 3D features. By combining depth, point cloud or voxel information, [39]–[41] achieve better 3D reconstruction via an encoder-decoder structure. [42], [43] create the mosaics via an online deep blending pipeline from multi-view stereo images. Dupont et al. [44] set a new path of neural rendering by enforcing equivariance between the change in viewpoint and change in the latent space.

*C. 3D-Aware View Synthesis*

After introducing 360° video/image processing and neural view rendering, let us further differentiate our proposed approach from other works: 3D-awareness is the core idea of different view synthesis approaches. In order to achieve better visual fidelity, the generative adversarial network is the most popular model, and has been successfully used in many works [19], [20], [24], [29]–[31]. Therefore, our method should be compared with three specific, key architectures: conditional GAN [23], HoloGAN [35], and ATSal [9].

As shown in Figure 2, we show the three architectures, where $T_3$ is the 3D camera pose, $T_2$ is 2D affine parameters, CMT is the cubemap projection (CMP) to project panorama into multiple views. Conditional GAN is a straightforward approach that takes the camera pose as the label (or condition) for image generation. The generator consists of 2D convolution layers. Since it does not allow explicit pose control, its capacity to generate multiple views is limited. HoloGAN was proposed to enable direct manipulation of viewing angles. This interesting architecture explicitly explores 3D representation of the training data. Given multi-view images of the same object, it uses 3D convolution to learn the 3D features of this object. It then applies 3D rigid-body transformations to the learned 3D features, and projects the transformed features to the 2D space for image generation.

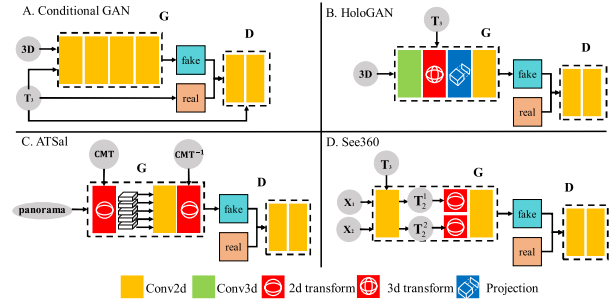

Fig. 2. Comparison of generative image models. G is the generator and D is the discriminator. "real" and "fake" are the ground truth and generated images. (a) and (b) are used for object-centered novel view rendering. (c) is used for salience prediction for 360° video. (d) is our proposed See360 model for view rendering. (a) Conditional GAN, it takes the target poses as label for training. (b) HoloGAN, it explicitly explores the camera pose by using directly the 3D rigid-body transformation. (c) ATSal, it uses cubemap projection (CMP) to project panorama into multiple views for further processing. (d) See360, it transfers 3D pose parameters to 2d affine parameters for view synthesis.

Though it achieves good performance on view synthesis, it is not suitable for 360° view synthesis, for two reasons. First, the 3D convolution is built on whatever the changes of view, the 3D representation remains the same to the given object. This works well for predicting different views of a given 3D object, but the method cannot predict new views for new 3D objects without some new training. Second, HoloGAN works well on simple objects with simple textures and no background, like chairs or faces. For complex scenes in real 3D worlds, it results in very blurry images [44]. ATSal is a model designed for 360° salience prediction. To find the spot where visual attention should focus, it processes panoramic, spherical videos by 2D convolution. It needs forward and backward cubemap projection to decompose the video at the different view and synthesize back the generated result. One of the problems is that such panoramas are not commonly used. They require specific collection, storage and processing mechanisms.

For comparison, we propose the See360 to achieve 360° view synthesis via a generative adversarial network. Instead of learning the complex 3D feature representation of the surroundings, we take two reference views and the target viewing angle to implicitly learn the 2D affine transformation which is equivalent to the new 3D view to be synthesized. Hence we tackle 3D view synthesis as a 2D image fusion problem with explicit pose control, and complement it with a model enabling to fill the missing/occluded regions.

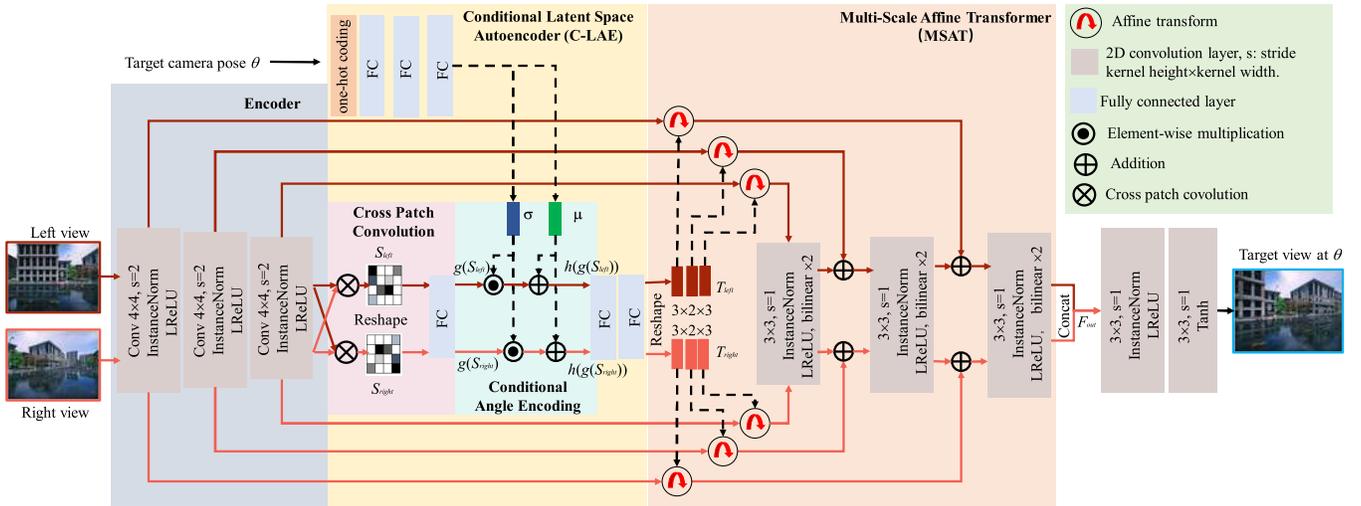

Fig. 3. The complete structure of the proposed generator of See360. It take left and right view as references to render a novel view with the target camera pose. It consists of 1) Encoder, Conditional Latent space AutoEncoder (C-LAE) and 2) Multi-Scale Affine Transformer (MSAT). Inside the C-LAE, there is a Cross Patch convolution to find view projection from left to right and from right to left, respectively. The target camera pose is injected for explicit view control via Conditional Angle Encoding. Then the affine transform is used in MSAT for reconstruction.

## III. METHOD

To render a novel view in a given camera pose, See360 extends traditional GANs by introducing a Conditional Latent space AutoEncoder (C-LAE) that maps the 3D camera pose to 2D image projection. It consists of 1) a novel cross-patch convolution and 2) an effective conditional angle encoding. Specifically, See360 fuses the two reference images by learning an equivalent 2D affine transform from their 3D camera poses, enabling the generated images to be a weighted interpolation of the two references.

To ensure the realistic visual quality of the novel view, See360 also includes a Multi-Scale Affine Transformer (MSAT) that renders edges and textures in a coarse-to-fine manner. Eventually, the generator will generate realistic views that can fool the discriminator. Similar to recent works [45]–[47], the GAN structure in See360 relies on a regular CNN as the discriminator, the key contribution being our new generator.

Figure 3 shows the complete structure of the generator of our proposed See360. It takes two references (left and right views) to render the novel view corresponding to the target camera pose $\theta$. The whole structure is built in U-shape to down- and up-scale the feature maps for feature extraction and fusion. For both left and right references, they share the same Encoder parameters for feature extraction. Then the left and right features are used for Cross patch convolution to learn implicit 3D feature representation. The Conditional Angle Encoding is used to take the camera pose code as input for pose manipulation. During the reconstruction, we use learned Affine matrices to transfer left and right features to the target camera pose. Finally, the transferred left and right features are concatenated to output the final image. We discuss each of these components in details in the following sections.

### A. Conditional Latent Space AutoEncoder

There are two key processes in our proposed Conditional Latent space AutoEncoder (C-LAE): 1) Cross Patch

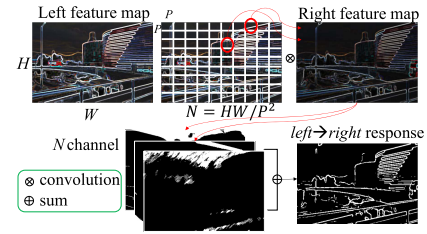

Fig. 4. Cross patch convolution from left to right features.

Convolution (the pink box in Figure 3) and 2) Conditional Angle Encoding (the cyan box in Figure 3). The former is for feature matching to learn the corresponding maps, while the latter is for camera pose manipulation.

*1) Cross Patch Convolution:* Given the input left and right views, there are 2D regions in the images that overlap in terms of content, and can thus be used as spatial clues to learn the hidden 3D correlation. In order to find the 2D overlapping regions, we use both left and right feature maps to search for pattern similarity. In other words, instead of looking for spatial correlation in the image space, we look for feature space similarity in a trainable convolution process.

To achieve this, as shown in Figure 4, we perform Cross patch convolution. We firstly split the left features $\mathbb{R}^{H \times W}$ into $N$ patches of size $P \times P$ and form a convolutional kernel $\mathbb{R}^{N \times P \times P}$, and the right features as target features for convolution. N convolution results are summed together to form the corresponding map from left-patches to the right features. Second, we apply the same process to find the right-patches that match the left features, this Cross patch convolution process can be expressed as:

$$left \rightarrow right \quad S^{left}_{i,j} = \sum_{a=0}^{h-1} \sum_{b=0}^{w-1} x_{a,b} \cdot Y_{i-a, j-b}$$

$$right \rightarrow left \quad S^{right}_{i,j} = \sum_{a=0}^{h-1} \sum_{b=0}^{w-1} y_{a,b} \cdot X_{i-a, j-b} \quad (1)$$

where $S_{i,j}^{left}$ and $S_{i,j}^{right}$ are the feature responses on location (i,j) from left to right view, and from right to left view, respectively. $X$ and $Y$ are the complete left and right features. $x$ and $y$ are the left and right feature patches of size $h \times w$. The output feature correspond map $S^{left}$ or $S^{right}$ is similar to an attention map that records the patch correspondence. It matches and attends the features from one view to another.

The use of the Cross patch convolution has two merits in our framework: 1) we can significantly reduce the feature dimension to find the key features in left and right views, and 2) we symmetrically apply Cross patch convolution for both left and right views so that we can find two independent feature correspondence maps.

*2) Conditional Angle Encoding:* After obtaining the two feature correspondence maps, we reshape them as 1D vectors and use one fully connected layer to find the implicit 3D representation as $g : S^i \to \Phi(S^i), i = (left, right)$. In order to have a flexible camera pose control, we design a conditional angle encoding scheme for the computation. Firstly, given left and right views with their camera pose interval (distance) $\tau$, we digitize the target camera pose as $\lfloor \frac{\theta}{\tau} \times \delta \rfloor$, where $\delta$ is the code length. Then we transfer the camera pose to a one-hot code as model input. For example, when we use $\delta = 13$ to digitize 60°, 30° can be coded as "0000001000000". Taking one-hot encoded target camera pose as input, we have three full connected layers to find the conditional vector defined by latent model as $f : z \to \mathcal{N}(\mu(z), \sigma(z))$, where z is the learned hidden vector via fully connection layers, $\mu(z)$ and $\sigma(z)$ are mean and variance for the hidden distribution. Similar to SFT [49], we treat the 3D-awareness as a "style" controller defined by the target camera pose:

$$h(g(S)) = g(S) * (1 + \sigma(z)) + \mu(z) \quad (2)$$

where $h(g(S))$ is the "normalized" vector that is adjusted by affine parameters for adaptive learning. Next, we use two fully connection layers to learn the equivalent 2D affine parameters $T_i \in \mathbb{R}^{2 \times 3}, i = (left, right)$. Empirically, we find this network architecture be able to disentangle pose much better than those that feed the camera pose directly to the first layer of the generator.

*B. Multi-Scale Affine Transformer*

For view generation, we propose to use a Multi-Scale Affine Transformer (MSAT) to reconstruct the image from coarse to fine scales. The idea is to parameterize the rigid-body transform by 2D rotation, scaling, shearing followed by bilinear resampling. As shown in Figure 5, the affine transformation is a simple image warping model which uses a six-degree parameter matrix T to describe the parametric planar transformation.

Note that we do not consider the translation in this work. Details on making this for pose sampling are discussed in Section IV. To adapt feature maps at different scales, instead of learning one uniform affine parameter for feature maps at different scales. In other words, the network is not designed for one-shot projection but multi-scale projection. Hence at each feature scale, we can finely adjust the projection to fuse

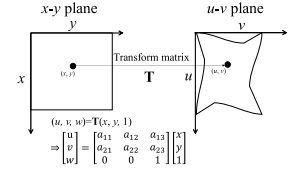

Fig. 5. Affine transformation.

them to the final result. Mathematically, we can describe the process as,

$$F_{out} = \sum_{j=1}^{3} Conv(T_{left}^j F_{left}^j \oplus T_{right}^j F_{right}^j) \quad (3)$$

where $F_{left}^j$ and $F_{right}^j$ are j-th encoder feature of left and right views, respectively. $T_{left}^j$ and $T_{right}^j$ are j-th affine matrices to transform features from left and right views. $\oplus$ is the concatenation operator to combine left and right features. Conv is the convolutional operation to upsample adjusted left and right features (gray boxes of MSAT in Figure 3). Two merits of performing multi-scale affine transform are: 1) more complex global perspective transform can be decomposed into several local affine transforms so that we can adaptively learn inlier features that match between two views using convolutional layers; 2) and performing an affine transform on multi-scale feature maps can correct subpixel misalignment in the spatial domain, which avoids stitching holes or scene distortion.

In Figure 3, we illustrate that we learn three sets of affine matrices ($3 \times 2 \times 3$) (which represents 3 sets of affine matrices of size $2 \times 3$) for left and right feature projection at different scales. Hence, we can first transform the left or right features, and then add them to the next level by upsampling $2\times$. Gradually, we upsample the feature map back to the original size. Finally, we concatenate transformed left and right feature maps to go through two convolution layers and fuse them into the final image. Eventually, we can obtain the novel view rendered at target camera pose $\theta$.

*C. Loss Functions*

Our new See360 model works as an image generator (G) to fuse two references into a novel view. To encourage the model to generate images with photo-realistic visual quality, we designed a simple 2D CNN network for multi-scale style discriminator (D) following the same idea as in [45], [46] to supervise the generation. Specifically, we use 2 discriminators ($D_1$, $D_2$) with the same network structure, but which operate at different image scales. Specifically, $D_1$ and $D_2$ respectively work on the prediction and ground truth images at downsampling factors of 1 and 2. They act as a minimax optimization as follows:

$$\min_G \max_{D_1, D_2} \sum_{k=1,2} \mathcal{L}(G(\theta, I^{left}, I^{right}),$$
$$D_k(I_\theta^{pre}, I^{left}, I^{right}, M_\theta^{gt})) \quad (4)$$

where $I^{left}$ and $I^{right}$ are the left and right references, $I_\theta^{pre}$ is the prediction at camera pose $\theta$. $M_\theta^{gt}$ is the ground truth segmentation map at camera pose $\theta$. In order to train a good discriminator to distinguish the prediction from the ground truth, we add three sources of inputs as conditions for training: i) left reference, ii) right reference and iii) semantic segmentation map. Hence the discriminator can learn to generate novel views with semantic 3D awareness with reference to left and right views and also the segmentation map.

*1) Generator Loss:* The generator is designed to match the image contents and style between the prediction and ground truth. For predicted $I_\theta^{pre}$ and ground truth $I_\theta^{gt}$ images at camera pose $\theta$, we use $\mathcal{L}_{ssim}(I_\theta^{pre}, I_\theta^{gt})$ to measure the Structural SIMilarity (SSIM) between the prediction and ground truth. To further encourage the structure similarity, we also use multiple Gaussian kernels to build Laplacian pyramid representation of image [51] to calculate the Laplacian loss $\mathcal{L}_{lap}(I_\theta^{pre}, I_\theta^{gt})$. Furthermore, we make use of VGG-19 to extract the deep feature representation of prediction and ground truth to estimate the distribution divergence $\mathcal{L}_{pd}$ [52] as follows.

$$\mathcal{L}_{pd} = \sum_{j=1} W_p(\phi(I_\theta^{pre}), \phi(I_\theta^{gt})) \quad (5)$$

where $W_p$ is the Wasserstein distance, $\phi(I_\theta^{pre})$ and $\phi(I_\theta^{gt})$ represent the feature maps obtained by the 4th convolution before the 5th maxpooling layer of VGG-19. To stabilize the training process, we also use $\mathcal{L}_{feat}$ [45] to compute the $L_1$ distance between the real and generated images using the intermediate features from discriminators by

$$\mathcal{L}_{feat} = \frac{1}{C}\sum_{j=1}^{C}\|D_{1,2}^j(I_\theta^{gt}, I^{left}, I^{right}, M_\theta^{gt}) \\ - D_{1,2}^j(I_\theta^{pre}, I^{left}, I^{right}, M_\theta^{gt})\| \quad (6)$$

where $C$ is the number of feature maps. The overall objective of the generator is to minimize the following loss function.

$$\mathcal{L}_G = \mathcal{L}_{adv}(G, D_{1,2}) + \lambda_{ssim}(1 - \mathcal{L}_{ssim}) \\ + \lambda_{pd}\mathcal{L}_{pd} + \lambda_{feat}\mathcal{L}_{feat} + \lambda_{lap}\mathcal{L}_{lap} \quad (7)$$

where $\mathcal{L}_{adv}$ is the conditional adversarial loss defined by

$$\mathcal{L}_{adv} = \frac{1}{2}\sum_{j=1}^{2} log[D_j(I_\theta^{gt}, I^{left}, I^{right}, M_\theta^{gt})] \\ + (1 - log[D_j(I_\theta^{pre}, I^{left}, I^{right}, M_\theta^{pre})]) \quad (8)$$

Since the generated scene should have the same semantic map as the ground truth scene, Equation 8 makes use of the segmentation map $M_\theta^{gt}$ (generated by pre-trained segmentation model) at camera pose $\theta$ as a condition for discriminator.

### D. Data Collection and Generation

Our target is to render 360° views for both indoor and outdoor environments. Most of the existing datasets are either collected as object-centred views from virtual machine [32] or indoor environment [33], [34]. To demonstrate the versatility and efficiency of the proposed See360 model, we propose two different types of datasets for training and evaluation: 1) virtual-world data, 2) real-world data. The basic process was to fix the camera at a given location. Then we horizontally rotated (along the y-axis) the camera to capture the images of 360° views. The angular step was set to 5°, enabling us to collect 72 views at each location. For each scene, we also collected the segmentation map. Note also that the datasets were captured to avoid many moving objects so that the static 3D structure can be learned by the proposed model.

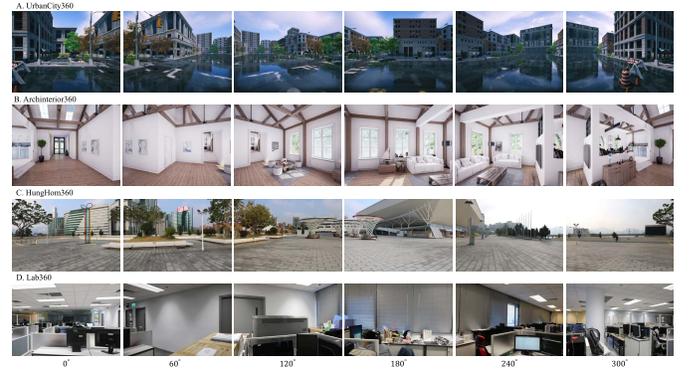

Fig. 6. Examples of our proposed UrbanCity360 (A), Archinterior360 (B), HungHom360 (C) and Lab360 (D) datasets. The UrbanCity360 and Archinterior360 are rendered from the virtual world. The HungHom360 and Lab360 are rendered from the real world.

For virtual-world data, we suggest using an open source project (unrealCV [50]) to render indoor and outdoor views from virtual worlds. Based on the available toolkit, we randomly placed the camera (distances range from 5m to 100m) and controlled it to capture images at different angles. For the indoor environment, we used the realistic virtual world "ArchinteriorsVol2Scene2" for data collection. It describes a scene of a house with 1 bedroom and 1 bathroom. We refer it to as "Archinterior360". For the outdoor environment, we used another realistic virtual world called "UrbanCity" for data collection, where the scene is a street block. We refer it to as "UrbanCity360". For "UrbanCity360" and "Archinterior360". We randomly placed the camera at 100 different locations to capture the scenes. We used another 10 locations to capture the images for evaluation. In summary, each dataset includes 7200 training images and 720 testing images. We also collect semantic segmentation maps from UnrealCV [50] for training.

For real-world data, as shown in Figure 1a and b, we installed the camera on an electronic tripod head to collect different views. We also collected both indoor and outdoor datasets for estimation (with distances ranging from 5m to 30m). For the outdoor dataset, we placed the camera on a street of Hung Hom, Hong Kong. We randomly chose 14 locations to collect a total of $14 \times 72 = 1008$ training images and another $5 \times 72 = 360$ images for testing. We refer this dataset to as "HungHom360". For the indoor dataset, we chose a laboratory in a university. We placed the camera at four corners of the lab to collect $4 \times 72 = 288$ training images and placed the camera at the center of a lab to collect 72 images for testing. We refer it to as "Lab360". We used pre-trained HRNet [53] to generate the corresponding segmentation maps for training (see equation 8).

As shown in Figure 4, the four datasets (UrbanCity360, Archinterior360, HungHom360 and Lab360) cover a large variety of indoor and outdoor environments with a large variety of contents. For example, UrbanCity360 contains views of a whole block of street. There are buildings, trees, street lamps and many other outdoor objects. More importantly, it has very strong lighting changes, such as shadows on the ground. Archinterior360 has more structural textures from the interior design, requiring a view rendering machine with a good 3D spatial awareness. HungHom360 contains natural lighting with a little fog. Because it was captured from an open area, the depth of the view changes a lot, which makes the rendering of novel views quite difficult. For Lab360, the challenge is that the ceiling lights cause uneven lighting conditions. Moreover, the number of small objects makes view prediction difficult.

## IV. EXPERIMENTS

### A. Datasets

Our goal being to achieve 360° neural view rendering without any prior about the 3D environment (e.g. no depth map, point cloud or other information), the left and right views we provide as input need to overlap slightly. In our experiments, we set the angle between them to 60° (which is also the FOV limit for most of the common cameras). This ensures limited overlap, enabling the images to show quite different scenes. To complete a 360° view, we can use 6 such references, to interpolate a maximum of 72-6 = 66 intermediate views. As introduced in Section III-A, we set the code length $\delta = 12$ so that every two references, we have 12 intermediate views, that is, the minimum angle change is $60°/12 = 5°$. All intermediate view images are the ones we used as ground truth for comparison. The datasets we used for evaluation are:

**UrbanCity360** and **Archinterior360** are virtual-world rendered datasets that consist of 7200 training images and 720 testing images of size $320 \times 240$.

**HungHom360** and **Lab360** are real-world rendered datasets with size $360 \times 240$. They were collected using DSLR.

### B. Implementation Details

*1) Network Architecture:* As shown in Figure 3, the generator of the proposed See360 method firstly downsamples the feature maps 3 times by a factor of 2. Each downsampling unit consists of 1 convolution layer (with kernel size $4 \times 4$, stride 2) followed by Instance Normalization and Leaky ReLU. The Conditional Latent space AutoEncoder consists of a few fully connected layers followed by ReLU activation functions. The Multi-Scale Affine Transformer upsamples the feature maps 3 times by a factor of 2 using simple bilinear interpolation. The structure of the discriminator is designed as 3 layers of convolution followed by Instance Normalization and Leaky ReLU. We trained our network using Adam optimizer with a learning rate of $10^{-4}$ and a batch size of 8. The training process for virtual-world datasets took about 8 hours on a single NVIDIA GTX1080Ti GPU. In contrast, the training on real-world datasets took about 10 mins at most because 1) we only have a few numbers of real-world images, so the training is fast and 2) we used the model pre-trained on virtual-world datasets as a starting point for fine-tuning, so we did not need too much training time. We conducted our experiments on a Pytorch platform. The source code and results will be made available upon publication of this work.

*2) Methods to Compare With:* This work being the first one, to our best knowledge, on camera-centered 360° neural view rendering, there is no existing approach tackling exactly the same problem that we can compare with. However, as shown in Section II, there are a few works that explored novel view rendering by proposing different network architectures. For comparison, we choose three state-of-the-art methods to implement their structures to resolve our problem: Conditional GAN [23], HoloGAN [35], Pix2pixHD [45] as shown in Figure 2. We followed the authors of these methods to design the networks, with a few modifications to achieve our goal. For Conditional GAN, we concatenated the left and right references as input, and the camera pose was directly introduced in the generator and discriminator as a condition for generation. For HoloGAN, we replaced the input 3D model with 2 reference images and added 4 layers of convolution to extract image features, then we followed the design in Figure 2B to add 3D convolution, projection and 2D convolution to generate images. For Pix2pixHD, we used the same network as the Conditional GAN with four modifications: 1) added Instance normalization on every convolution layer, 2) added $\mathcal{L}_{feat}$ to stabilize the training process, 3) used multi-scale style discriminators to train the whole GAN network and 4) introduced segmentation map as one extra condition for the generator. We also compared with a classic panorama approach by using open software Hugin [65]. For all methods, we reused the original codes provided by the authors and trained them using the suggested settings, including the same programming platform, training iterations, parameter initialization and training strategy, to ensure network optimization. For a fair comparison, we trained and tested all methods using the same datasets.[1]

### C. Evaluation Metrics

*1) Quantitative Evaluation:* To objectively evaluate the data fidelity of view rendering, we have used PSNR, SSIM and running time. To avoid the boundary effect, the computation of PSNR and SSIM excludes a region of 8 pixels wide around the image boundaries.

*2) Qualitative Evaluation:* To better evaluate the visual quality of view rendering than just visualizing the generated images, we used LPIPS [54] to measure the deep feature similarity. It is widely used in image processing tasks [55]–[57]. Using HRNet [53], we also estimated the semantic segmentation map from the generated image to compare with the ground truth by averaging pixel accuracy (%).

### D. Comparisons With Prior Works

Let us compare our approach and other methods on 4 different datasets, using the 5 different evaluation metrics just discussed. Results are shown in Table I. As illustrated in

---

[1]Note that panorama cannot synthesize new view given specific camera poses, what we did is to use given views to form 360 panorama and then uniformly crop subimages from the panorama as intermediate views.

TABLE I

QUANTITATIVE COMPARISON OF THE PROPOSED SEE360 AND OTHER NOVEL VIEW RENDERING NETWORKS ON 4 DATASETS, INCLUDING PSNR (dB), SSIM, LPIPS, PIXEL ACCURATE(%) AND RUNNING TIME (s/IMG). FOR PSNR, SSIM AND PIXEL ACCURACY, HIGHER VALUES MEAN BETTER RESULTS. FOR LPIPS AND RUNNING TIME, LOWER VALUES MEAN BETTER RESULTS

| Virtual-world datasets | UrbanCity360 | | | | | Archinterior360 | | | | |
|---|---|---|---|---|---|---|---|---|---|---|
| | PSNR | SSIM | LPIPS | Pixel Accuracy | Time | PSNR | SSIM | LPIPS | Pixel Accuracy | Time |
| Hugin [65] | 18.08 | 0.502 | 0.379 | 0.690 | 1.102 | 17.12 | 0.601 | 0.499 | 0.732 | 1.102 |
| Conditional GAN [23] | 17.92 | 0.436 | 0.383 | 0.621 | 0.299 | 16.50 | 0.531 | 0.476 | 0.654 | 0.311 |
| HoloGAN [35] | 17.56 | 0.388 | 0.458 | 0.611 | 0.287 | 15.80 | 0.418 | 0.499 | 0.561 | 0.271 |
| Pix2pixHD [45] | 18.10 | 0.451 | 0.351 | 0.702 | 0.321 | 16.70 | 0.533 | 0.454 | 0.699 | 0.351 |
| See360 (Ours) | 25.49 | 0.727 | 0.142 | 0.800 | 0.351 | 22.99 | 0.706 | 0.165 | 0.832 | 0.319 |
| Real-world datasets | HungHom360 | | | | | Lab360 | | | | |
| | PSNR | SSIM | LPIPS | Pixel Accuracy | Time | PSNR | SSIM | LPIPS | Pixel Accuracy | Time |
| Hugin [45] | 16.19 | 0.402 | 0.519 | 0.711 | 1.011 | 15.01 | 0.551 | 0.604 | 0.688 | 1.011 |
| Conditional GAN [23] | 15.31 | 0.337 | 0.466 | 0.608 | 0.279 | 14.44 | 0.429 | 0.578 | 0.630 | 0.288 |
| HoloGAN [35] | 15.13 | 0.331 | 0.513 | 0.502 | 0.312 | 13.88 | 0.409 | 0.577 | 0.542 | 0.311 |
| Pix2pixHD [45] | 15.39 | 0.337 | 0.464 | 0.620 | 0.297 | 14.49 | 0.429 | 0.582 | 0.633 | 0.300 |
| See360 (Ours) | 21.06 | 0.603 | 0.161 | 0.797 | 0.335 | 22.06 | 0.744 | 0.133 | 0.764 | 0.337 |

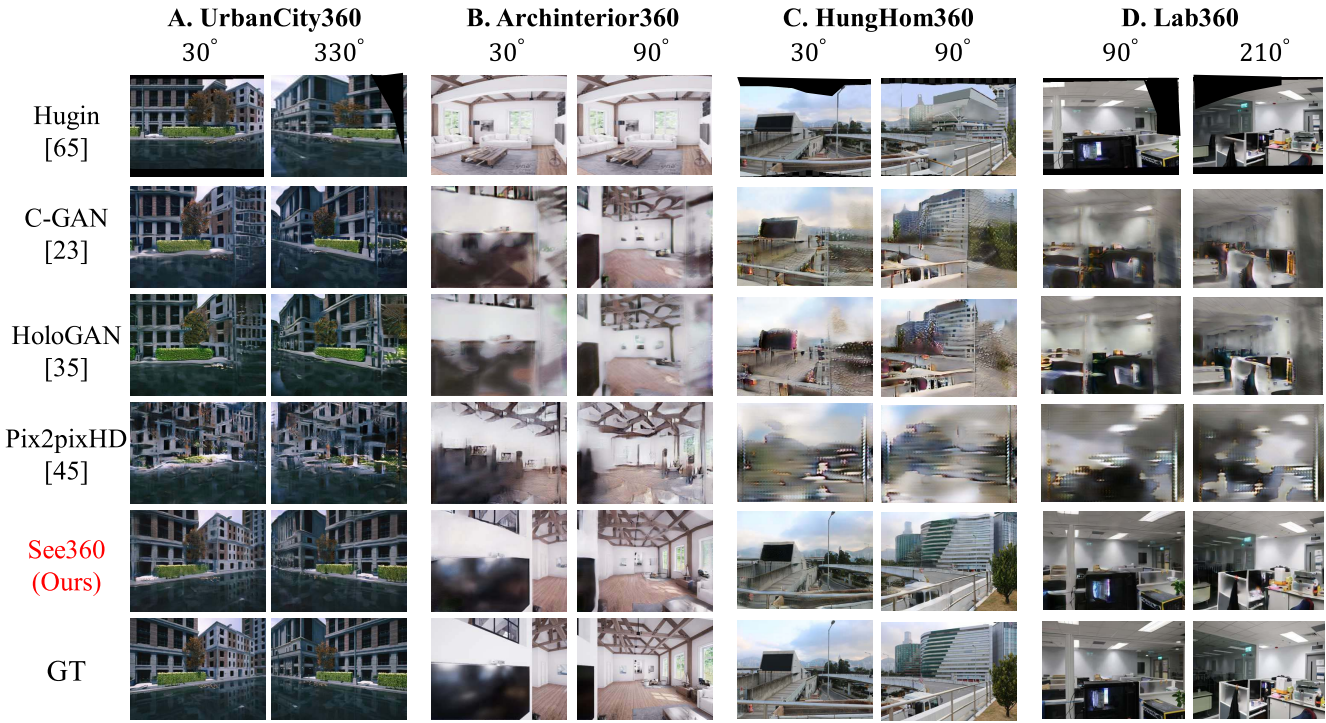

Fig. 7. Visual comparison among different methods on virtual-world and real-world datasets.

Section III-B, we sparsely selected 6 references (0°, 60°, 120°, 180°, 240° and 300°) to cover the whole 360° view. For each two adjacent references, every 5° angle, we predicted one view leading to the estimation of 6 × 12.72 novel views, totally. We used PSNR and SSIM to evaluate data fidelity compared to ground truth. We found that our proposed See360 achieves the best performance compared to others by about 0.6 dB and 0.3, respectively. For LPIPS, we used the pre-trained AlexNet [54] to measure feature distance, which describes the semantic differences on the deep feature representation. We observed that See360 also achieves the best performance by about 0.2 in terms of LPIPS values. For pixel accuracy, we used the pre-trained semantic segmentation network [53] to estimate the pixel-wise accuracy on the layout of the image. It can classify over 100 objects, like trees, buildings and so on. We can also see that our approach outperforms others by about 2%. We also list the computational time to compare the computation complexity of the different methods. This shows that the four methods have similar running times (with ±0.04s differences), so See360 is not more costly than others. Without adding any complex computation or processing, its quality still outperforms other methods by a large margin for three reasons: 1) our proposed See360 is tailored for novel view synthesis without using 3D information, 2) other methods are proposed for limited view synthesis, i.e., HoloGAN, C-GAN and Pix2pixHD are proposed for camera-centered view synthesis so that they do not fit the target of 360° view rendering, and 3) other methods focus on synthetic/simple objects rather than real-world scenes, hence they do no perform well.

For visual comparison, Figure 7 shows two examples from virtual-world datasets and two examples from real-world datasets (where C-GAN stands for Conditional GAN). We visualize both the generated images as well as the segmentation maps. In Figure 7A, we can see that the proposed See360 can reconstruct the details of the buildings, while others distorted the textures. For example, though the 30° views predicted by C-GAN and Pix2pixHD look alike real scenes, they fail to predict the correct angles of the scene. In Figure 7B, we can see clear differences among the different methods. The challenge of this scene is that the lighting condition is vividly rendered by the virtual machine and the interior structure is highly correlated. C-GAN, Pix2pixHD and HoloGAN all fail to reconstruct the scenes. Hugin's results, on the other hand, have the problem of filling the holes of the new scenes. In addition, from the example for 30° and 90° of Archinterior360, it fails to either fill the whole scene or predict the unseen contents. In contrast, See360 can clearly reconstruct the wooden ceiling and windows. For real-world examples in Figure 7C and D, we can see that C-GAN, Pix2pixHD and HoloGAN fail to adjust their models to these real scenes. Hugin still has the problem of filling the holes of the new scenes. It even misses the building from the 90° view of HungHom360. Our proposed See360 can still predict the real scenes (outdoor or indoor) at different viewing angles. For example, in Figure 7C, on the view at 210°, See360 can well predict the trees on the left and the colorful statue on the right. In Figure 7D, See360 does not only estimate the workplace, but also renders the lighting conditions, such as generating shadows cast by the lights.

These results show that our proposed See360 method successfully achieves 360° view synthesis, on both virtual-world and real-world scenes. In contrast, the other three methods fail. For example, we can consider that the Conditional GAN and Pix2pixHD can resolve the problem to some extent but do not provide the necessary 3D awareness to render from a specific viewing angle. The reason is that Conditional GAN and Pix2pixHD use the camera pose as an extra input of the network to implicitly control the view rendering. On the contrary, our model explicitly ports the camera pose into the network as 2D affine transformation on the feature space. For HoloGAN, it does not work on different datasets because it uses 3D convolution to explicitly discover the 3D representation. However, the real scene is not object centered. Since views from different angles describe different 3D structures, 3D convolution cannot uniformly project different scenes to a single 3D space for estimation. In contrast, our proposed See360 method learns to use equivalent 2D affine transform to estimate the 3D model for novel views of different angles.

### E. Ablation Study

In the ablation study, we consider the effect of three key components: 1) Cross patch correlation (CPC), 2) Multi-Scale

TABLE II

QUANTITATIVE COMPARISON OF THE PROPOSED SEE360 WITH OR WITHOUT FOUR COMPONENTS ON URBANCITY360 AND HUNGHOM360 DATASETS, INCLUDING PSNR (DB), SSIM AND LPIPS

| Eval | UrbanCity360 | | | HungHom360 | | |
|---|---|---|---|---|---|---|
| | PSNR | SSIM | LPIPS | PSNR | SSIM | LPIPS |
| w/o CPC | 25.12 | 0.720 | 0.188 | 20.56 | 0.598 | 0.203 |
| | (-0.37) | (-0.007) | (-0.046) | (-0.50) | (-0.005) | (-0.042) |
| w/o MSAT | 22.87 | 0.659 | 0.257 | 18.96 | 0.589 | 0.263 |
| | (-2.62) | (-0.068) | (-0.115) | (-0.210) | (-0.014) | (-0.102) |
| w/o SS | 25.35 | 0.724 | 0.166 | 20.78 | 0.600 | 0.189 |
| | (-0.14) | (-0.003) | (-0.024) | (-0.28) | (-0.003) | (-0.028) |
| Full pipeline | 25.49 | 0.727 | 0.142 | 21.06 | 0.603 | 0.161 |

Affine Transformer (MSAT) and 3) multi-scale style discriminator with or without semantic segmentation (SS). To make a comparison, we refer to these three situations to as CPC, MSAT and SS. In Table II, we compute the evaluation metrics PSNR, SSIM and LPIPS on each dataset, to quantitatively measure the importance of their effects.

In Table II, "Full Pipeline" is the our complete See360 model. To test the three key components, we successively removed the key components from the full pipeline to train the model. We can find that the most important component is the use of Multi-Scale Affine Transformer (MSAT). The reason is that we can reconstruct the scene from coarse to fine to fill the details at different scales. Meanwhile, the proposed MSAT adaptively estimates different 2D affine parameters for features at different scales to fuse the reference views. Similar to "CPC", we use it to extract the global feature correspondence. It can also improve the performance by about 0.3 dB and 0.01 in terms of PSNR and SSIM, respectively. For "SS", we consider it as an extra constraint on the image generation. As discussed in Equation 8, using semantic segmentation as layout information (the contour of the whole scene) and combining it with generated images, the discriminator can learn to distinguish whether the generated image has similar structural information as ground truth.

In Table III, we did another ablation study using different losses introduced in Section III-C. The key losses we discussed include Laplacian loss (*lap loss*), Projection Distribution loss (*pd loss*) and VGG feature loss (*VGG loss*). *See360—* indicates models using different combinations of losses and *See360* indicates the best model that makes use of all three losses. We measured PSNR, SSIM and LPIPS on UrbanCity360 dataset. It can be found that using *lap loss* improves the SSIM because it computes structural similarity of images using the differences of Gaussians. Using *pd loss* and *VGG loss* improves the LPIPS but affects PSNR and SSIM scores because of the trade-off effect between pixel distortion and visual quality [66]. Combining all three loss terms, *See360* achieves the best LPIPS as well as the second best PSNR and SSIM, which indicates that our proposed See360 can well synthesize photo-realistic novel views.

### F. Robust and Flexible 3D Awareness

As discussed in Section II, our proposed See360 method can recognize different camera poses to synthesize unique proposed See360 could predict novel views every 5°. The two examples in Figure 8 further demonstrate the flexible 3D awareness of our method. They show that the proposed See360 model can recognize every camera pose to render novel scenes with different viewing angles. The visual quality of predicted views is good enough that can be used for view interpolation. It inspires us that it has the potential to be used for 360° scenery video. We provide a video demo at the link https://youtu.be/P1JHx7ViSpI.[2] Please note the smooth motion transition and view changes. There could be some blurry effects or distortions due to sudden big structural changes or additional noises on the reference images. Visual quality can be further improved by using larger datasets and image pre-processing.

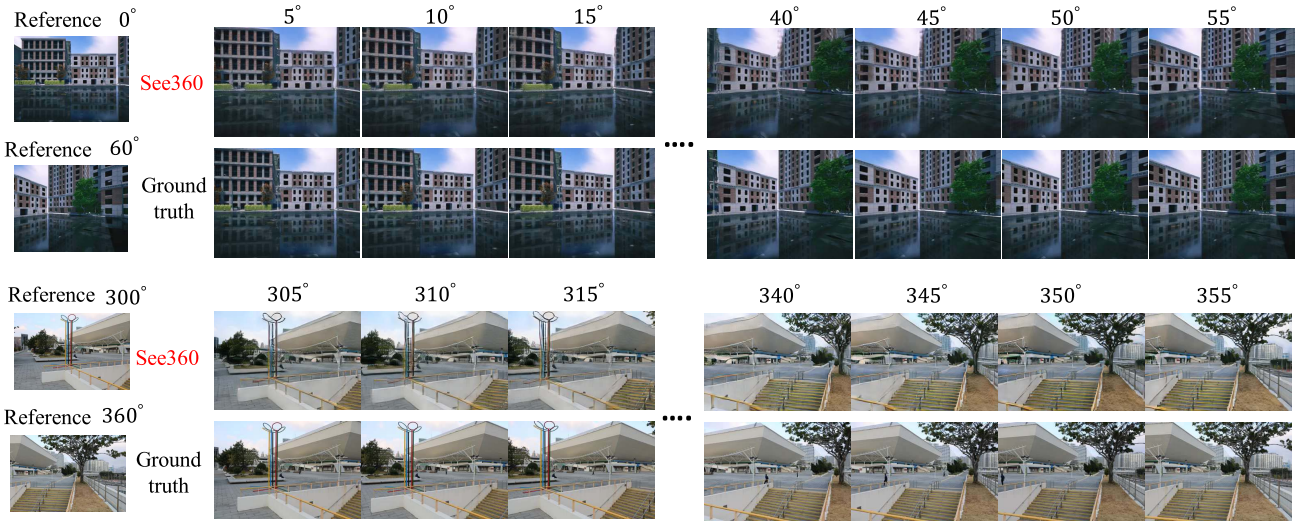

Fig. 8. Visualization of view synthesis of fine camera poses on UrbanCity360 and HungHom360 datasets.

TABLE III
QUANTITATIVE COMPARISON OF THE PROPOSED SEE360 USING DIFFERENT LOSS TERMS, INCLUDING LAPALACIAN LOSS (*lap loss*), PROJECTION DISTRIBUTION LOSS (*pd loss*) AND VGG FEATURE LOSS (*VGG loss*)

| Model | lap loss | pd loss | VGG loss | PSNR | SSIM | LPIPS |
|---|---|---|---|---|---|---|
| | | | | 25.47 | 0.727 | 0.479 |
| | ✓ | | | 25.62 | 0.747 | 0.395 |
| See360— | | ✓ | | 25.38 | 0.699 | 0.1750 |
| | | | ✓ | 25.37 | 0.682 | 0.169 |
| | ✓ | | ✓ | 25.45 | 0.719 | 0.194 |
| | | ✓ | ✓ | 25.44 | 0.719 | 0.192 |
| See360 | ✓ | ✓ | ✓ | 25.49 | 0.727 | 0.142 |

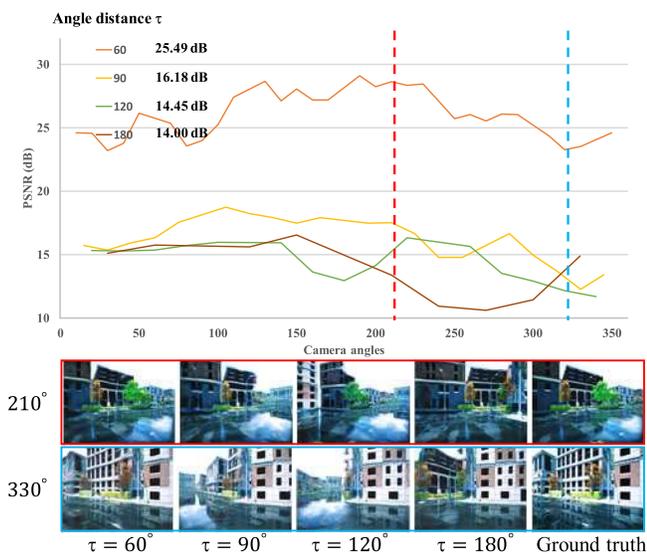

Fig. 9. Visualization of view synthesis using references with different angle distances. The upper figure shows the PSNR results against the camera pose. Different colors of the lines represent the different references for testing. On the lower side, we show two predicted views at camera pose 210° and 330°. The average PSNR is listed on the right top corner.

To test the limit of the See360 model, we also investigated whether it can use references with a larger angle distance, while still predicting the novel view correctly. We used UrbanCity360 as an example to show the results of using references with different $\tau$ in Figure 9, rather than always using $\tau = 60°$. Note that the model was still trained using references with $\tau = 60°$. During testing, we deliberately used references with different angle distances ($\tau = 60°$, $90°$, $120°$ or $180°$) to predict the intermediate views. We reported the average PSNR on the right top corner. It can be seen that using $\tau = 60°$ achieves the best results in terms of PSNR because it suits the setting of the training. As expected, with the angle distance becoming larger, the less scene overlap between references, so worse was our prediction. However, the proposed See360 can still be able to fill the missing scenes with similar patterns. For instance, if we check the predicted view using $\tau = 90°$, our model still generates a good result.

Another aspect of robust 3D awareness is the real-time rendering ability for practical applications. That is, we directly use our pre-trained model on new scenes without any fine-tuning. We collected real-world datasets to evaluate such views. In our setting for training, the largest angle interval (distance) between left and right references was 60° and our versatility for the proposed See360 model. Since real-world datasets contain much fewer images for training, we firstly pre-trained the model on virtual-world datasets and then fine-tuned them on the targeted real-world datasets for better prediction. As shown in Table I, we only fine-tuned our models for 100 epochs to adjust to real-world datasets, which required about 10 mins on a computer with one GPU. Figure 10 illustrates the robustness of See360 to different datasets. To further illustrate the robustness, we show the intermediate results on fine-tuning See360, using the HungHom360 dataset. With the training epochs increase, we can see that the visual quality is getting better, e.g. see the stairs and trees. In addition, we can find that even when we trained the model for 5 epochs, it can already achieve reasonable results.

[2]We highly recommend the readers to check the video for high-quality video comparison since some figures in the paper are much smaller than our actual view predictions.

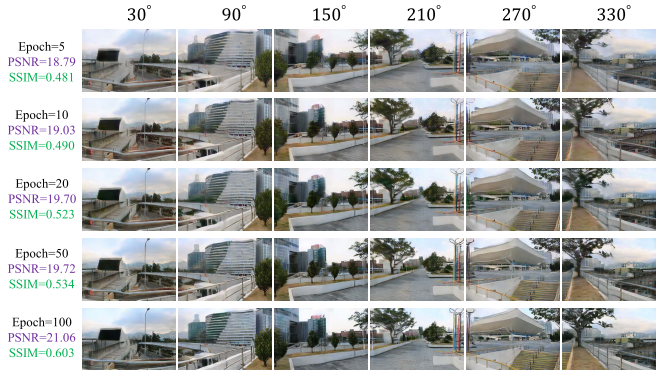

Fig. 10. Visualization of view synthesis on HungHom360 dataset. We demonstrate the intermediate training results at different epochs.

### G. Novel View Synthesis in the Wild

So far, we presented evaluations on the few datasets on which the model was trained. In addition, the proposed model works well in the wild. For example, given two reference views of an unknown scene, we do not even need to know the angle between these left and right references: We represent each intermediate view as a unique one-hot code, so all we need is the one-hot code of the target camera pose. To illustrate this, we collected a few views inside a university by randomly rotating the camera and we used any two views as references to interpolate the middle view. E.g., we input the one-hot camera pose code as "0000001000000" for prediction. Note again that we do not train on unknown scenes.

We directly use the pre-trained UrbanCity360 and Archinterior360 models for prediction. As shown in Figure 11, though UrbanCity360 and Archinterior360 are virtual-world datasets, we can find that both models can render novels view with similar structures. Another interesting finding is that using UrbanCity360 model can achieve better results than Archinterior360. The possible reason is that UrbanCity360 is also an outdoor dataset with diverse contents including buildings and trees. Therefore, it is a better match for unknown outdoor scenes.

However, when we have photos with views that highly differ from our training data, like photos taken from other countries, the proposed See360 model may not work properly. Since we cannot actually go around different places to test different views, as shown in Figure 12, we used Google map to locate three landmarks in Paris, London and New York, as Louvre museum, British museum and Manhattan street. Google map can provide panorama views, so we captured the panorama and cut it into three sub-images to mimic the multi-view image generation. We took the left and right images as references to predict the center view. We directly applied the proposed See360 (learned from our training data) to these three scenes. It shows that See360 can predict some of the views but with some artifacts. However, there are clear directions for improvement, for two reasons: 1) the panorama consists of a wide view computed from several blended images, so it contains discontinuous lens distortion. Therefore, the extracted left and right references images do not meet the criteria of our setting for training. 2) The captured views are from around the world, therefore are very different from the views we used for training. We used these challenging examples to show that the architecture of See360 is solid and it can be further improved when larger datasets are used for training.

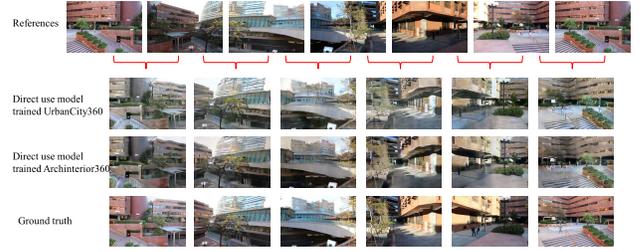

Fig. 11. Visualization of view synthesis on unknown scenes captured inside a campus. We do not train the model but directly use two pre-trained models for estimation: 1) pre-trained UrbanCity360 model and 2) pre-trained Archinterior360 model.

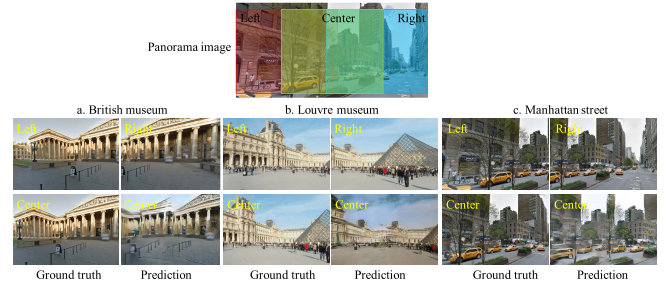

Fig. 12. Novel view synthesis of unknown scenes captured around the world, including Louvre museum in Paris, British museum in London and Manhattan street in New York. The left and right views are cut from a panorama, and the center view is used for prediction.

### V. CONCLUSION

In this paper, we open up a new research direction of using neural rendering to generate views at different viewing angles from only two overlapping input images, a key contribution helping people understand their surroundings. Our novel view synthesis method, called See360, builds on two carefully designed components: 1) Multi-Scale Affine Transformer (MSAT) and 2) Conditional Latent space AutoEncoder (C-LAE). The key insight is that we transfer 3D view rendering as an equivalent 2D affine transformation. We contribute further by providing two types of datasets, respectively consisting of synthetic and real images, for training and evaluation. Our See360 model has the potential to be used in 360° video processing for virtual reality. In future work, we plan to explore the extension of See360 to combine both 2D and 3D information for high-resolution images/videos rendering.

## ACKNOWLEDGMENT

The authors thank Dr. Li-Wen Wang for helping data collection and equipment setting.

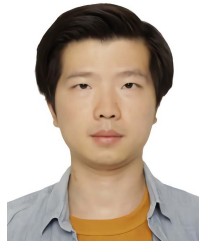

**Zhi-Song Liu** (Member, IEEE) received the Ph.D. degree in electronic engineering from The Hong Kong Polytechnic University, Hong Kong, in 2020. He was with École Polytechnique in 2020. He is currently a Postdoctoral Fellow with the Caritas Institute of Higher Education. He has published over 16 articles in journals, such as IEEE TRANSACTIONS ON IMAGE PROCESSING and IEEE TRANSACTIONS ON MULTIMEDIA, and conferences, such as ICIP, CVPR, ECCV, and ICCV. His research interests include image and video signal processing, computer vision, deep learning, and 3D data processing.

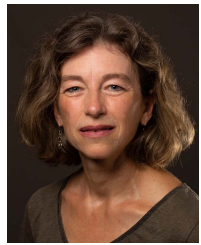

**Marie-Paule Cani** is currently a Professor of computer science at École Polytechnique. Her research interests cover both shape modeling and computer animation. She contributed over the years to a number of high level models for shapes and motion, such as implicit surfaces, multi-resolution physically-based animation methods, and hybrid representations for real-time natural scenes. Following a long lasting interest for virtual sculpture, she has been recently searching for more expressive ways to create 3D contents, such as combining sketch-based interfaces with procedural models based on a priori knowledge or learning. She received the Eurographics Outstanding Technical Contributions Award in 2011 and the Silver Medal from CNRS in 2012, and was elected at the Academia Europaea in 2013. She was awarded the ERC Advanced Grant EXPRESSIVE (2012–2017) and joined the ACM Siggraph Academy in 2019. She was elected at the French Academy of Sciences in 2020.

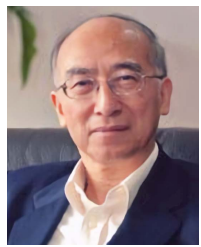

**Wan-Chi Siu** (Life Fellow, IEEE) received the M.Phil. degree from The Chinese University of Hong Kong in 1977 and the Ph.D. degree from Imperial College London in 1984. He was a Chair Professor and the Head of the EIE Department and the Dean of the Engineering Faculty, The Hong Kong Polytechnic University. He is now an Emeritus Professor with The Hong Kong Polytechnic University and a Research Professor of the Caritas Institute of Higher Education. He is an expert in DSP, transforms, fast algorithms, machine learning, deep learning, super-resolution imaging, 2D and 3D video coding, object recognition, and tracking. He has published 500 research papers (over 200 appeared in international journals), edited three books, and nine recent patents granted. He was the Independent Non-Executive Director (2000–2015) of a publicly-listed video surveillance company and a Convenor of the First Engineering/IT Panel of the RAE (1992–1993), Hong Kong. He is a fellow of IET. He is an outstanding scholar, with many awards, including the Best Teacher Award, the Best Faculty Researcher Award (twice), and the IEEE Third Millennium Medal (2000). He has been a Guest Editor/Subject Editor/AE of IEEE TRANSACTIONS ON CIRCUITS AND SYSTEMS—II: EXPRESS BRIEFS, IEEE TRANSACTIONS ON IMAGE PROCESSING, IEEE TRANSACTIONS ON CIRCUITS AND SYSTEMS FOR VIDEO TECHNOLOGY, and *Electronics Letters*, and organized very successfully over 20 international conferences, including IEEE society-sponsored flagship conferences, such as the TPC Chair of ISCAS1997 and the General Chair of ICASSP2003 and ICIP2010. He was the President (2017–2018) of Asia–Pacific Signal and Information Processing Association, and the Vice-President, the Chair of Conference Board, and a Core Member of Board of Governors (2012–2014) of the IEEE Signal Processing Society. He has been a member of the IEEE Educational Activities Board, the IEEE Fourier Award for Signal Processing Committee (2017–2020), and some other IEEE technical committees.